\documentclass{article} 
\usepackage{iclr2024_conference,times}

\usepackage{amsmath,amsfonts,bm}

\def\eqref#1{equation~\ref{#1}}

\def\1{\bm{1}}

\DeclareMathAlphabet{\mathsfit}{\encodingdefault}{\sfdefault}{m}{sl}
\SetMathAlphabet{\mathsfit}{bold}{\encodingdefault}{\sfdefault}{bx}{n}

\DeclareMathOperator*{\argmax}{arg\,max}

\usepackage{hyperref}
\usepackage{url}

\usepackage{graphicx}
\usepackage{booktabs}
\usepackage{dsfont}
\usepackage{algorithm}
\usepackage{algpseudocode}
\usepackage{array}
\usepackage{multirow}
\usepackage{paralist}
\usepackage{diagbox}
\usepackage{tikz}
\usetikzlibrary{tikzmark,calc}
\usepackage[leftcaption]{sidecap}
\graphicspath{ {./} }
\usepackage{refcount}
\usepackage{wrapfig}

\usepackage{tablefootnote}
\newcommand{\footnoteagain}[1]{\hyperref[#1]{\footnotemark[\getrefnumber{#1}]}}
\usepackage{pifont}

\title{Ensemble Distillation for\\Unsupervised Constituency Parsing}

\author{Behzad Shayegh$^{1,*}$ \;\; Yanshuai Cao$^2$ \;\; Xiaodan Zhu$^{3,4}$ \;\; Jackie C.K. Cheung$^{5,6}$ \;\; Lili Mou$^{1,6}$
\\
$^1$Dept. Computing Science, Alberta Machine Intelligence Institute (Amii), University of Alberta\\
$^2$Borealis AI\quad
$^3$Dept. Electrical and Computer Engineering, Queen’s University\\
$^4$Ingenuity Labs Research Institute, Queen’s University\\
$^5$Quebec Artificial Intelligence Institute (MILA), McGill University\quad
$^6$Canada CIFAR AI Chair\\
\texttt{\href{mailto:the.shayegh@gmail.com}{the.shayegh@gmail.com}} \\
\texttt{\href{mailto:yanshuai.cao@borealisai.com}{yanshuai.cao@borealisai.com}}\qquad
\texttt{\href{mailto:xiaodan.zhu@queensu.ca}{xiaodan.zhu@queensu.ca}} \\
\texttt{\href{mailto:jcheung@cs.mcgill.ca}{jcheung@cs.mcgill.ca}} \qquad\qquad\qquad
\texttt{\href{mailto:doublepower.mou@gmail.com}{doublepower.mou@gmail.com}}
}

\iclrfinalcopy 
\begin{document}

\maketitle

\begingroup
\renewcommand\thefootnote{$\star$}
\footnotetext{Work partially done during Mitacs internship at Borealis AI.}
\endgroup

\begin{abstract}
We investigate the unsupervised constituency parsing task, which organizes words and phrases of a sentence into a hierarchical structure without using linguistically annotated data. We observe that existing unsupervised parsers capture different aspects of parsing structures, which can be leveraged to enhance unsupervised parsing performance.
To this end, we propose a notion of ``tree averaging,'' based on which we further propose a novel ensemble method for unsupervised parsing.
To improve inference efficiency, we further distill the ensemble knowledge into a student model; such an ensemble-then-distill process is an effective approach to mitigate the over-smoothing problem existing in common multi-teacher distilling methods.
Experiments show that our method surpasses all previous approaches, consistently demonstrating its effectiveness and robustness across various runs, with different ensemble components, and under domain-shift conditions.\footnote{Code available at \url{https://github.com/MANGA-UOFA/ED4UCP}}
\end{abstract}

\section{Introduction}
\label{sec:introduction}

Constituency parsing is a well-established task in natural language processing (NLP), which interprets a sentence and induces its constituency tree, a syntactic structure representation that organizes words and phrases into a hierarchy~\citep{constituencyTree}. It has wide applications in various downstream tasks, including semantic role labeling~\citep{mohammadshahi-henderson-2023-syntax} and explainability of AI models~\citep{tenney-etal-2019-bert,wu2022weakly}.
Traditionally, parsing is accomplished by supervised models trained with linguistically annotated treebanks~\citep{charniak-2000-maximum}, which are expensive to obtain and may not be available for low-resource scenarios.
Also, these supervised parsers often underperform when encountering domain shifts. 
This motivates researchers to explore unsupervised methods as they eliminate the need for annotated data.

To address unsupervised parsing, researchers have proposed various heuristics and indirect supervision signals. \citet{clark-2001-unsupervised} employs context distribution clustering to induce a probabilistic context-free grammar~\citep[PCFG;][]{Booth1969ProbabilisticRO}. \citet{klein-manning-2002-generative} define a joint distribution for sentences and parse structures, the latter learned by expectation--maximization (EM) algorithms. \citet{snyder2009unsupervised} further extend unsupervised parsing to the multilingual setting with bilingual supervision.

In the deep learning era, unsupervised parsing techniques keep advancing. \citet{cao-etal-2020-unsupervised} utilize linguistic constituency tests~\citep{constituencyTree} as heuristics, evaluating all spans as potential constituents for selection. \citet{li-lu-2023-contextual} modify each span based on linguistic perturbations and observe changes in the contextual representations of a masked language model; according to the level of distortion, they determine how likely the span is a constituent. \citet{maveli-cohen-2022-co} use rules to train two classifiers with local features and contextual features, respectively, which are further refined in a co-training fashion. Another way to obtain the parsing structure in an unsupervised way is to treat it as a latent variable and train it in downstream tasks, such as text classification~\citep{li-etal-2019-imitation}, language modeling~\citep{shen2018ordered,kim-etal-2019-unsupervised}, and sentence reconstruction~\citep{drozdov-etal-2019-unsupervised-latent,kim-etal-2019-compound}. Overall, unsupervised parsing is made feasible by such heuristics and indirect supervisions, and has become a curious research direction in NLP.

In our work, we uncover an intriguing phenomenon of low correlation among different unsupervised parsers, despite their similar overall $F_1$ scores (the main evaluation metric for parsing), shown in Table~\ref{tab:correlation-table}.
While \citet{williams-etal-2018-latent} have shown low self-consistency in early latent-tree models, we go further and show the correlation among different models is even lower than restarts of the same model. This suggests that existing unsupervised parsers capture different aspects of the structures, and our insight is that combining these parsers may leverage their different expertise to achieve higher performance for unsupervised parsing.

To this end, we propose an ensemble method for unsupervised parsing. We first introduce a notion of ``tree averaging'' based on the similarity of two constituency trees. Given a few existing unsupervised parsers, referred to as \textit{teachers},\footnote{Our full approach involves training a student model from the ensemble; thus, it is appropriate to use the term \textit{teacher} for an ensemble component.} we then propose to use a CYK-like algorithm~\citep{Kasami1965AnER,YOUNGER1967189,MANACHER1978127,sennrich-2014-cyk} that utilizes dynamic programming to search for the tree that is most similar to all teachers' outputs. In this way, we are able to obtain an ``average'' parse tree, taking advantage of different existing unsupervised parsers.

To improve the inference efficiency, we distill our ensemble knowledge into a student model. In particular, we choose the recurrent neural network grammar~\citep[RNNG;][]{dyer-etal-2016-recurrent} with an unsupervised self-training procedure~\citep[URNNG;][]{kim-etal-2019-unsupervised}, following the common practice in unsupervised parsing~\citep{kim-etal-2019-compound,cao-etal-2020-unsupervised}. Our ensemble-then-distill process is able to mitigate the over-smoothing problem, where the standard cross-entropy loss encourages the student to learn an overly smooth distribution \citep{wen-etal-2023-f}. Such a problem exists in common multi-teacher distilling methods~\citep{wu-etal-2021-one}, and would be especially severe when the teachers are heterogeneous, signifying the importance of our approach.

We evaluated our ensemble method on the Penn Treebank~\citep[PTB;][]{marcus-etal-1993-building} and SUSANNE~\citep{SusanneCorpus} corpora. Results show that our approach outperforms existing unsupervised parsers by a large margin in terms of $F_1$ scores, and that it achieves results comparable to the supervised counterpart in the domain-shift setting. Overall, our paper largely bridges the gap between supervised and unsupervised constituency parsing.

In short, the main contributions of this paper include: 1) We reveal an intriguing phenomenon that existing unsupervised parsers have diverse expertise, which may be leveraged by model ensembles; 2) We propose a notion of tree averaging and utilize a CYK-like algorithm that searches for the average tree of existing unsupervised parsers; and 3) We propose an ensemble-then-distill approach to improve inference efficiency and to alleviate the over-smoothing problem in common multi-teacher distilling approaches.

\begin{table}[t]
\vspace{-9pt}
    \begin{center}
        \resizebox{0.9\linewidth}{!}{
            \begin{tabular}{c c c c}
                \cline{2-2}
                \multicolumn{1}{r|}{Compound PCFG} & \multicolumn{1}{c|}{$100$} & & \\
                \cline{2-3}
                \multicolumn{1}{r|}{DIORA} & \multicolumn{1}{c|}{$55.8$} & \multicolumn{1}{c|}{$100$} & \\
                \cline{2-4}
                \multicolumn{1}{r|}{S-DIORA} & \multicolumn{1}{c|}{$58.1$} & \multicolumn{1}{c|}{$63.6$} & \multicolumn{1}{c|}{$100$} \\
                \cline{2-4}
                &Compound & DIORA & S-DIORA \\
                &PCFG &  & \\
            \end{tabular}
            \quad
            \begin{tabular}{c c c c}
                \cline{2-2}
                \multicolumn{1}{r|}{DIORA\ding{73}} & \multicolumn{1}{c|}{$100$} & & \\
                \cline{2-3}
                \multicolumn{1}{r|}{DIORA\ding{71}} & \multicolumn{1}{c|}{$74.1$} & \multicolumn{1}{c|}{$100$} & \\
                \cline{2-4}
                \multicolumn{1}{r|}{DIORA\ding{91}} & \multicolumn{1}{c|}{$74.3$} & \multicolumn{1}{c|}{$74.9$} & \multicolumn{1}{c|}{$100$} \\
                \cline{2-4}
                & DIORA\ding{73} & DIORA\ding{71} & DIORA\ding{91} \\
            \end{tabular}
        }
    \end{center}\vspace{-8pt}
    \caption{Correlation analysis of unsupervised parsers. Numbers are the $F_1$ score of one model against another on the Penn Treebank dataset~\citep{marcus-etal-1993-building}. The left table considers three heterogeneous models (Compound PCFG, DIORA, and S-DIORA), whereas the right table considers three runs (\ding{73}, \ding{71}, and \ding{91}) of the same model. All their $F_1$ scores against the groundtruth fall within the range of 59--61, thus providing a controlled experimental setting.}
    \label{tab:correlation-table}
\end{table}

\section{Approach}

\subsection{Unsupervised Constituency Parsing}
\label{subsection:unsupConstPars}

In linguistics, a \textit{constituent} refers to a word or a group of words that function as a single unit in a hierarchical tree structure~\citep{constituencyTree}. In the sentence ``The quick brown fox jumps over the lazy dog,'' for example, the phrase ``the lazy dog'' serves as a noun phrase constituent, whereas ``jumps over the lazy dog'' is a verb phrase constituent. In this study, we address \textit{unsupervised} constituency parsing, where no linguistic annotations are used for training. This reduces human labor and is potentially useful for low-resource languages. Following most previous work in this direction~\citep{cao-etal-2020-unsupervised,maveli-cohen-2022-co,li-lu-2023-contextual}, we focus on \textit{unlabeled, binary} parse trees, in which each constituent has a binary branching and is not labeled with its syntactic role (such as a noun phrase or a verb phrase).

The standard evaluation metric for constituency parsing is the $F_1$ score, which is the harmonic mean of precision and recall~\citep{shen2017neural,SupervisedCRFConstituencyParsing}:
\begin{align}P = \frac{|C(T_{\text{pred}}) \cap C(T_{\text{ref}})|}{|C(T_{\text{pred}})|},\;\; R = \frac{|C(T_{\text{pred}}) \cap C(T_{\text{ref}})|}{|C(T_{\text{ref}})|},\;\; F_1 = 2 \frac{PR}{P+R}
\end{align}
where $T_{\text{pred}}$ and $T_{\text{ref}}$ are predicted and reference trees, respectively, and $C(T)$ is the set of constituents in a tree $T$.

\subsection{A Notion of Averaging Constituency Trees} 

In our study, we propose an ensemble approach to combine the expertise of existing unsupervised parsers (called \textit{teachers}), as we observe they have low correlation among themselves despite their similar overall $F_1$ scores (Table~\ref{tab:correlation-table}).

To accomplish ensemble binary constituency parsing, we need to define a notion of tree averaging; that is, our ensemble output is the average tree that is most similar to all teachers' outputs. Inspired by the evaluation metric, we suggest the average tree should have the highest total $F_1$ score compared with different teachers. Let $s$ be a sentence and $T_k$ be the $k$th teacher parser. Given $K$ teachers, we define the average tree to be
\begin{align}
\label{eq:AvgTree}
\operatorname{AvgTree}(s, \{T_k\}_{k=1}^K) = \argmax\limits_{T \in \mathcal T(s)} \sum_{k=1}^K F_1(T, T_k(s))
\end{align}
where $\mathcal T(s)$ is all possible unlabeled binary trees on sentence $s$, and $T_k(s)$ is the $k$th teacher's output. 

It is emphasized that only the trees of the same sentence can be averaged. This simplifies the $F_1$ score of binary trees, as the denominators for both precision and recall are $2|s|-1$ for a sentence with $|s|$ words, i.e., $|C(T_{\text{pred}})| = |C(T_{\text{ref}})| = 2|s|-1$. Thus, Eqn.~(\ref{eq:AvgTree}) can be rewritten as:
\begin{align}
\operatorname{AvgTree}(s, \{T_k\}_{k=1}^K) &= \argmax\limits_{T \in \mathcal T(s)} \sum_{k=1}^K F_1(T, T_k(s)) = \argmax\limits_{T \in \mathcal T(s)} \sum_{k=1}^K \frac{|C(T) \cap C(T_k(s))|}{2|s|-1} \\
\label{eq:SimpleAvgTree}
&= \argmax\limits_{T \in \mathcal T(s)} \sum_{c \in C(T)} \underbrace{\sum_{k=1}^K \mathds{1}[c \in C(T_k(s))]}_{\operatorname{HitCount}(c, \{T_k(s)\}_{k=1}^K)}
\end{align}
Here, we define the $\operatorname{HitCount}$ function to be the number of times that a constituent $c$ appears in the teachers’ outputs. In other words, Eqn.~(\ref{eq:SimpleAvgTree}) suggests that the average tree should be the one that hits the teachers' predicted constituents most.

\textbf{Discussion on MBR decoding.}
\label{discussion:SelectiveMBR}
Our work can be seen as minimum Bayes risk (MBR) decoding~\citep{MBRbook}. In general, MBR yields an output that minimizes an \textit{expected error} (called \textit{Bayes risk}), defined according to the task of interest. In our case, the error function can be thought of as $-\sum_{c\in C(T)}\operatorname{HitCount}(c, \{T_k(s)\}_{k=1}^K)$, and minimizing such-defined Bayes risk is equivalent to maximizing the total hit count in Eqn.~(\ref{eq:SimpleAvgTree}).

However, our MBR approach significantly differs from prior MBR studies in NLP. In fact, MBR has been widely applied to text generation~\citep{kumar-byrne-2004-minimum,10.1162/tacl_a_00491,suzgun-etal-2023-follow}, where a set of candidate output sentences are obtained by sampling or beam search, and the best one is selected based on a given error function, e.g., the dissimilarity against others; such an MBR method is \textit{selective}, meaning that the output can only be selected from a candidate set. On the contrary, our MBR is \textit{generative}, as the sentence's entire parse space $\mathcal T(s)$ will be considered during the $\argmax$ process in (\ref{eq:SimpleAvgTree}).
This follows \citet{petrov-klein-2007-improved} who search for the global lowest-risk tree in the task of supervised constituency parsing.
Here, the global search is feasible because the nature of tree structures facilitates efficient exact decoding with dynamic programming, discussed in the next subsection.

\subsection{Our CYK Variant}\label{Sec:OurCYK}

As indicated by Eqn.~(\ref{eq:SimpleAvgTree}), our method searches for the constituency tree with the highest total hit count of its constituents in the teachers' outputs. We can achieve this by a CYK~\citep{Kasami1965AnER,YOUNGER1967189}-like dynamic programming algorithm, because an optimal constituency parse structure of a \textit{span}---a continuous subsequence of a sentence---is independent of the rest of the sentence.

Given a sentence $s$, we denote by $s_{b:e}$ a span starting from the $b$th word and ending right before the $e$th word. Considering a set of teachers $\{T_k\}_{k=1}^K$, we define a recursion variable 
\begin{align}\label{eq:H}
H_{b:e}=\max_{T\in\mathcal T({s_{b:e}})}\sum_{c\in C(T)}\operatorname{HitCount}(c, \{T_k(s)\}_{k=1}^K)
\end{align}
which is the best total hit count for this span.\footnote{Note that, in Eqns.~(\ref{eq:H})--(\ref{eq:CTbuilding}), $T_k(s)$ should not be $T_k(s_{b:e})$, because the hit count is based on the teachers' sentence-level parsing.} We also define $L_{b:e}$ to be the corresponding, locally best parse structure, unambiguously represented by the set of all constituents in it.

The base cases are $H_{b:b+1} = K$ and $L_{b:b+1}=\{s_{b:b+1}\}$ for $b=1,\cdots, |s|$, suggesting that the best parse tree of a single-word span is the word itself, which appears in all teachers' outputs and has a hit count of $K$.

For recursion, we see a span $s_{b:e}$ will be split into two subspans $s_{b:j}$ and $s_{j:e}$ for some split position~$j$, because our work focuses on binary constituency parsing. Given $j$, the total hit count for the span $s_{b:e}$ is the summation over those of the two subspans $s_{b:j}$ and $s_{j:e}$, plus its own hit count. To obtain the best split, we need to vary $j$ from $b$ to $e$ (exclusive), given by
\begin{align}
\label{eq:splitpointJ}
j_{b:e}^* = \argmax\limits_{b < j < e} 
\left[H_{b:j}+H_{j:e}+\operatorname{HitCount}(s_{b:e}, \{T_k(s)\}_{k=1}^K)\right]
\end{align}
where the hit count is a constant for $\argmax$ and can be omitted in implementation. Then, we have
\begin{align}
H_{b:e} &= H_{b:j_{b:e}^*} + H_{j_{b:e}^*:e} + \operatorname{HitCount}(s_{b:e}, \{T_k(s)\}_{k=1}^K)\\
\label{eq:CTbuilding}
L_{b:e} &= L_{b:j_{b:e}^*} \cup L_{j_{b:e}^*:e} \cup \{s_{b:e}\}
\end{align}
Eqn.~(\ref{eq:CTbuilding}) essentially groups two sub-parse structures $L_{b:j_{b:e}^*}$ and $L_{j_{b:e}^*:e}$ for the span $s_{b:e}$. This can be represented as the union operation on the sets of constituents.

The recursion terminates when we have computed $L_{1:|s|+1}$, which is the best parse tree for the sentence $s$, maximizing overall similarity to the teachers' predictions and being our ensemble output.
In Appendix~\ref{apdx:CYKalgo}, we summarize our ensemble procedure in pseudocode and provide an illustration. 

\subsection{Ensemble Distillation}
\label{sec:ensembledistillation}

In our work, we further propose an ensemble distilling approach that trains a \textit{student} parser from an ensemble of teachers. This is motivated by the fact that the ensemble requires performing inference for all teacher models and may be slow.
Specifically, we choose the recurrent neural network grammar~\citep[RNNG;][]{dyer-etal-2016-recurrent} as the student model, which learns shift--reduce parsing operations~\citep{shiftreduce} along with language modeling using recurrent neural networks.  The choice of RNNG is due to its unsupervised refinement procedure~\citep[URNNG;][]{kim-etal-2019-unsupervised}, which treats syntactic structures as latent variables and uses variational inference to optimize the joint probability of syntax and language modeling, given some unlabeled text. Such a self-training process enables URNNG to significantly boost parsing performance. 

Concretely, we treat the ensemble outputs as pseudo-groundtruth parse trees and use them to train RNNG with cross-entropy loss. Then, we apply URNNG for refinement, following previous work~\citep{kim-etal-2019-compound,cao-etal-2020-unsupervised}.

\textbf{Discussion on union distillation.}
\label{discussion:uniondistill}
An alternative way of distilling knowledge from multiple teachers is to perform cross-entropy training based on the union of the teachers' outputs~\citep{wu-etal-2021-one}, which we call \textit{union distillation}. Specifically, the cross-entropy loss between a target distribution~$t$ and a learned distribution $p$ is $-\sum_x t(x)\log p(x)$, which tends to suffer from an over-smoothing problem~\citep{WeiDialogGenerateShort,wen2022equal,wen-etal-2023-f}: the machine learning model will predict an overly smooth distribution $p$ to cover the support of $t$; if otherwise $p(x)$ is zero but $t(x)$ is non-zero for some~$x$, the cross-entropy loss would be infinity. Such an over-smoothing problem is especially severe in our scenario, as will be shown in Section~\ref{ss:analysis}, because our multiple teachers are heterogeneous and have different expertise (Table~\ref{tab:correlation-table}). By contrast, our proposed method is an ensemble-then-distill approach, which first synthesizes a best parse tree by model ensemble and then learns from the single best tree given an input sentence.

\section{Experiments}

\subsection{Datasets}

We evaluated our approach on the widely used Penn Treebank~\citep[PTB;][]{marcus-etal-1993-building} dataset, following most previous work~\citep{shen2018ordered,kim-etal-2019-compound,cao-etal-2020-unsupervised,maveli-cohen-2022-co,li-lu-2023-contextual}. We adopted the standard split: 39,701 samples in Sections 02--21 for training, 1,690 samples in Section 22 for validation, and 2,412 samples in Section 23 for test. It is emphasized that we did not use linguistic annotations in the training set, but took the unlabeled sentences to train teacher unsupervised parsers and to distill knowledge into the student.

In addition, we used the SUSANNE dataset~\citep{SusanneCorpus} to evaluate model performance in a domain-shift setting. Since it is a small, test-only dataset containing 6,424 samples in total, it is not possible to train unsupervised parsers directly on SUSANNE, which on the other hand provides an ideal opportunity for domain-shift evaluation.

We adopted the standard evaluation metric, the $F_1$ score of unlabeled constituents, as has been explained in Section~\ref{subsection:unsupConstPars}. We used the same evaluation setup as \citet{kim-etal-2019-compound}, ignoring punctuation and trivial constituents, i.e., single words and the whole sentence. We reported the average of sentence-level $F_1$ scores over the corpus.

\subsection{Competing Methods}

Our ensemble approach involves the following classic or state-of-the-art unsupervised parsers as our teachers, which are also baselines for comparison.

\begin{compactitem}[\quad$\bullet$]
\item \textbf{Ordered Neurons}~\citep{shen2018ordered}, a neural language model that learns syntactic structures with a gated attention mechanism;

\item \textbf{Neural PCFG}~\citep{kim-etal-2019-compound}, which utilizes neural networks to learn a latent probabilistic context-free grammar;

\item \textbf{Compound PCFG}~\citep{kim-etal-2019-compound}, which improves the Neural PCFG by adding an additional sentence-level latent representation;

\item \textbf{DIORA}~\citep{drozdov-etal-2019-unsupervised-latent}, a deep inside--outside recursive auto-encoder that 
marginalizes latent parse structures during encoder--decoder training;

\item \textbf{S-DIORA}~\citep{drozdov2020diora}, a variant of DIORA that only considers the single most probable tree during unsupervised training;

\item \textbf{ConTest}~\citep{cao-etal-2020-unsupervised}, which induces parse trees by rules and heuristics inspired by constituency tests~\citep{mccawley1998syntactic}; and

\item \textbf{ContexDistort}~\citep{li-lu-2023-contextual}, which induces parsing structures from pretrained masked language models---in particular, the BERT-base model~\citep{devlin-etal-2019-bert} in our experiments---based on contextual representation changes caused by linguistic perturbations.
\end{compactitem}

To combine multiple teachers, we consider several alternatives:
\begin{compactitem}[\quad$\bullet$]
\item \textbf{Selective MBR}, which selects the lowest-risk constituency tree among a given candidate set (Section~\ref{discussion:SelectiveMBR}). In particular, we consider teachers' outputs as the candidates, and we have     $\operatorname{SelectiveMBR}(s, \{T_k\}_{k=1}^K) = \argmax_{T \in \mathcal  \{T_k(s)\}_{k=1}^K} \sum_{k=1}^K F_1(T, T_k(s))$. This differs from our MBR approach, which is generative and performs the $\argmax$ over the entire binary tree space, shown in Eqn.~(\ref{eq:AvgTree}).

\item \textbf{Union distillation}, which trains a student from the union of the teachers' outputs (Section~\ref{sec:ensembledistillation}).
\end{compactitem}

For hyperparameters and other setups of previous methods (all teacher and student models), we used default values mentioned in either papers or codebases. It should be emphasized that our proposed ensemble approach does not have any hyperparameters, thus not requiring any tuning.

\subsection{Main Results}

\textbf{Results on PTB.} Table~\ref{tab:main_results} presents the main results on the PTB dataset, where we performed five runs of replication either by loading original authors' checkpoints or by rerunning released codebases. Our replication results are comparable to those reported in previous papers, inventoried in Appendix~\ref{apdx:ExperimentDetail}, showing that we have successfully established a foundation for our ensemble research.

We first evaluate our ensembles of corresponding runs (Row~12), which is a fair comparison against teacher models (Rows~3--9). Without RNNG/URNNG distillation, our method outperforms the best teacher (Row~8) by $7.5$ points in terms of $F_1$ scores, showing that our ensemble approach is highly effective and justifying the proposed notion of tree averaging for unsupervised parsing. 

It is also possible to have an ensemble of the best (or worst) teachers, one per each model across different runs, as the teacher models are all validated by a labeled development set. We observe that the ensemble of the best (or worst) teachers achieves slightly higher (or lower) scores than the ensemble of the teachers in corresponding runs, which is intuitive. However, the gap between the best-teachers ensemble and worst-teachers ensemble is small (Rows~13 vs.~14), showing that our approach is not sensitive to the variance of teachers. Interestingly, the ensemble of the worst teachers still outperforms the best single teacher by a large margin of $6.5$ $F_1$ points. 

We compare our ensemble approach with selective MBR (Row~11), which selects a minimum-risk tree from the teachers' predictions. As shown, selective MBR outperforms all the teachers too, again verifying the effectiveness of our tree-averaging formulation. However, its performance is worse than our method (Row~12), which can be thought of as generative MBR that searches the entire tree space using a CYK-like algorithm.

\begin{table}[t]
\centering
\resizebox{0.87\linewidth}{!}{
\begin{tabular}{r l | l | >{\;} l l l}
\toprule
& Method & Mean$_{\pm\text{Std}}$ & Run~1 & +RNNG & +URNNG\\
\midrule
1& Left branching & $\ \ 8.7$ & -- & -- & --\\
2& Right branching & $39.5$ & -- & -- & --\\
\midrule
3& Ordered Neurons~\citep{shen2018ordered} & $44.3_{\pm6.0}$ & $44.8$ & $45.4$ & $45.3$\\
4& Neural PCFG~\citep{kim-etal-2019-compound} & $51.0_{\pm1.7}$ & $48.4$ & $48.9$ & $51.1$\\
5& Compound PCFG~\citep{kim-etal-2019-compound} & $55.5_{\pm2.4}$ & $60.1$ & $60.5$ & $67.4$\\
6& DIORA~\citep{drozdov-etal-2019-unsupervised-latent} & $58.9_{\pm1.8}$ & $55.4$ & $58.6$ & $62.3$\\
7& S-DIORA~\citep{drozdov2020diora} & $57.0_{\pm2.1}$ & $56.3$ & $59.4$ & $62.2$\\
8& ConTest~\citep{cao-etal-2020-unsupervised} & $62.9_{\pm1.6}$ & $65.9$ & $64.6$ & $68.5$\\
9& ContexDistort~\citep{li-lu-2023-contextual} & $47.8_{\pm0.9}$ & $48.8$ & $48.5$ & $50.8$\\
\midrule
10& Union distillation  & -- & -- & $65.6$ & $65.4$\\
11& Selective MBR & $66.3_{\pm0.6}$ & $66.7$ & $68.6$ & $71.5$\\
12& Our ensemble (corresponding run) & $70.4_{\pm0.6}$ & $70.5$ & $69.7$ & $71.7$\\
13& Our ensemble (worst teacher across runs) & $69.4$ & -- & $69.1$ & $70.0$\\
14& Our ensemble (best teacher across runs) & $\textbf{71.9}$ & -- & $\textbf{71.1}$ & $\textbf{72.8}$\\
\midrule
15& Oracle & $83.3$ & -- & $76.0$ & $76.0$ \\
\bottomrule
\end{tabular}
}
\caption{$F_1$ scores on PTB. Teacher models' results are given by our five runs of replication (detailed in Appendix~\ref{apdx:ExperimentDetail}) for a fair comparison. 
Due to the limit of computing resources, we trained RNNG/URNNG with the first run only. The oracle refers to the highest possible $F_1$ score of a binary tree, as the groundtruth tree may not be binary.
}
\label{tab:main_results}
\end{table}

Then, we evaluate the distillation stage of our approach, which is based on Run~1 of each model. We observe our RNNG and URRNG follow the same trend as in previous work that RNNG may slightly hurt the performance, but its URNNG refinement\footnote{URNNG is traditionally used as a refinement procedure following a noisily trained RNNG~\citep{kim-etal-2019-compound, cao-etal-2020-unsupervised}. If URNNG is trained from scratch, it does not yield meaningful performance and may be even worse than right-branching~\citep{kim-etal-2019-unsupervised}. Thus, we excluded URNNG from our teachers.} yields a performance boost. 
It is also noted that URNNG's boosting effect on our approach is less significant than that on previous models, which is reasonable because our ensemble (w/o RNNG or URNNG) has already achieved a high performance. Overall, we achieve an $F_1$ score of $72.8$ in Row~14, being a new state of the art of unsupervised parsing and largely bridging the gap between supervised and unsupervised constituency parsing.

We compare our ensemble-then-distill approach with union distillation (Row~10), which trains from the union of the teachers' first-run outputs. As expected in Section~\ref{discussion:uniondistill}, union distillation does not work well; its performance is worse than that of the best teacher (Run~1 of Row~8), suggesting that multiple teachers may confuse the student and hurt the performance. Rather, our approach requires all teachers to negotiate a most agreed tree, thus avoiding confusion during the distilling process.

\begin{wraptable}{r}{0.45\textwidth}
\centering
\resizebox{\linewidth}{!}{
\begin{tabular}{r l | l l l}
\toprule
& Method & Run 1 & +RNNG & +URNNG\\
\midrule
1& Left branching & $\ \ 6.9$ & -- & --\\
2& Right branching & $26.9$ & -- & --\\
\midrule
3& Ordered Neurons & $32.4$ & $33.1$ & $33.1$\\
4& Neural PCFG & $44.2$ & $46.1$ & $48.6$\\
5& Compound PCFG & $43.0$ & $43.4$ & $46.5$\\
6& DIORA & $35.9$ & $42.2$ & $44.0$\\
7& S-DIORA & $37.5$ & $43.3$ & $42.4$\\
8& ConTest & $38.8$ & $46.9$ & $47.3$\\
9& ContexDistort & $41.2$ & $39.7$ & $41.1$\\
\midrule
10& Selective MBR & $47.4$ & $48.4$ & $48.5$\\
11& Our ensemble &$\textbf{50.3}$ & $49.1$ & $48.8$\\
\midrule
12& PTB-supervised & -- & $50.1$ & $49.8$\\
13& SUSANNE oracle & $69.8$ & -- & --\\
\bottomrule
\end{tabular}
}\vspace{-7pt}
\caption{$F_1$ scores in the domain-shift setting from PTB to SUSANNE. Note that all models were trained on PTB, including RNNGs and URNNGs. Since our approach is highly robust, we only considered the models of the first run on PTB in this experiment.}\vspace{-10pt}
\label{tab:susanne_results}
\end{wraptable}

\textbf{Results on SUSANNE.} Table~\ref{tab:susanne_results} presents parsing performance under a domain shift from PTB to SUSANNE. We directly ran unsupervised PTB-trained models on the test-only SUSANNE corpus without finetuning. This is a realistic experiment to examine the models' performance in an unseen low-resource domain.

We see both selective MBR (Row~10) and our method (Row~11) outperform all teachers (Rows~3--9) in the domain-shift setting, and that our approach outperforms selective MBR by $3$ points. The results are consistent with the PTB experiment.

For the ensemble-distilled RNNG and URNNG (Row~11), the performance drops slightly, probably because the performance of our ensemble approach without RNNG/URNNG is saturating and close to the PTB-supervised model (Row~12), whose RNNG/URNNG distillation also yields slight performance drop.
Nevertheless, our RNNG and URNNG (Row~11) outperform all the baselines in all settings. Moreover, the inference of our student model does not require querying the teachers, and is $18$x faster than the ensemble method (Appendix~\ref{app:efficiency}). Thus, the ensemble-distilled model is useful as it achieves competitive performance and high efficiency.

\subsection{In-Depth Analysis}
\label{ss:analysis}

\textbf{Denoising vs.~utilizing different expertise.} A curious question raised from the main results is why our ensemble approach yields such a substantial improvement. We have two plausible hypotheses: 1) The ensemble approach merely smooths out the teachers' noise, and 2) The ensemble approach is able to utilize different expertise of heterogeneous teachers.

We conducted the following experiment to verify the above hypotheses. Specifically, we compare two settings: the ensemble of three runs of the same model and the ensemble of three heterogeneous models. We picked the runs and models such that the two settings have similar performance. This sets up a controlled experiment, as the gain obtained by the ensemble of multiple runs suggests a denoising effect, whereas a further gain obtained by the ensemble of heterogeneous models suggests the effect of utilizing different expertise.

\begin{figure}[t]\vspace{-20pt}
\begin{center}
\includegraphics[width=\linewidth]{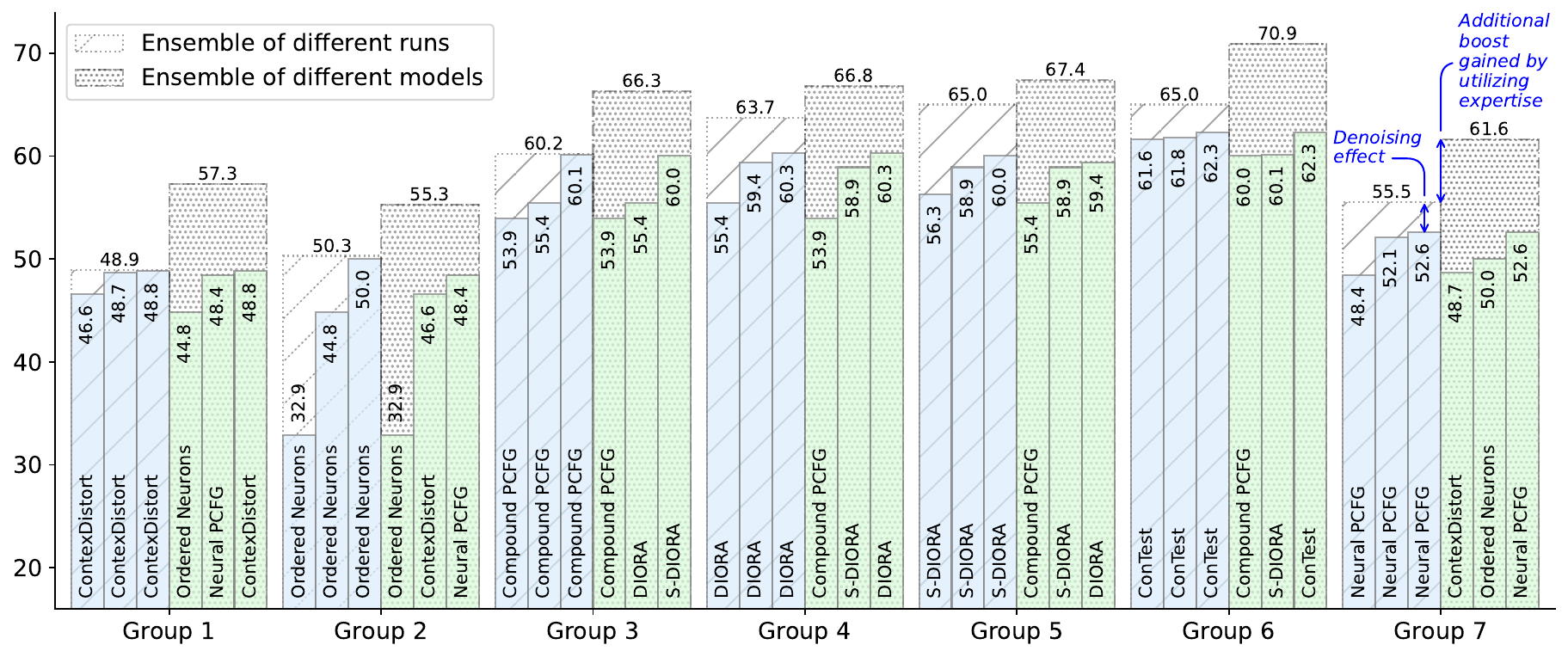}
\end{center}\vspace{-11pt}
\caption{Effect of denoising vs.~utilizing different expertise. Results are the $F_1$ scores on the PTB test set. The {\color{blue} \textit{italic blue annotation}} is an interpretation of the plot.}
\label{fig:denoisingvsexp}
\end{figure}

We repeated the experiment for seven groups with different choices of models and show results in Figure~\ref{fig:denoisingvsexp}. As seen, the ensemble of different runs always outperforms a single run, showing that the denoising effect does play a role in the ensemble process. Moreover, the ensemble of heterogeneous models consistently leads to a large add-on improvement compared with the ensemble of multiple runs; the results convincingly verify that different unsupervised parsers learn different aspects of the language structures, and that our ensemble approach is able to utilize such different expertise.

\textbf{Over-smoothing in multi-teacher knowledge distillation.} As discussed in Section~\ref{discussion:uniondistill}, union distillation is prone to the over-smoothing problem, where the student learns an overly smooth, wide-spreading distribution. This is especially severe in our setting, as our student learns from multiple heterogeneous teachers.

\begin{table}[t]\vspace{-3pt}
\centering
\resizebox{.5\linewidth}{!}{
\begin{tabular}{lcc}
\toprule
Approach & Mean Entropy & Std \\
\midrule
Union distillation    & $11.42$       & $0.09$              \\
Our ensemble distillation & $4.93$        & $0.12$              \\
Binarized-groundtruth supervision & $2.26$        & $0.12$              \\
\bottomrule
\end{tabular}
}\vspace{-5pt}
\caption{The mean and standard deviation of the prediction entropy for distilled/supervised RNNGs.}
\label{tab:Entropy}\vspace{-5pt}
\end{table}

\begin{wrapfigure}{r}{0.45\textwidth}
\vspace{-15pt}
    \includegraphics[width=\linewidth]{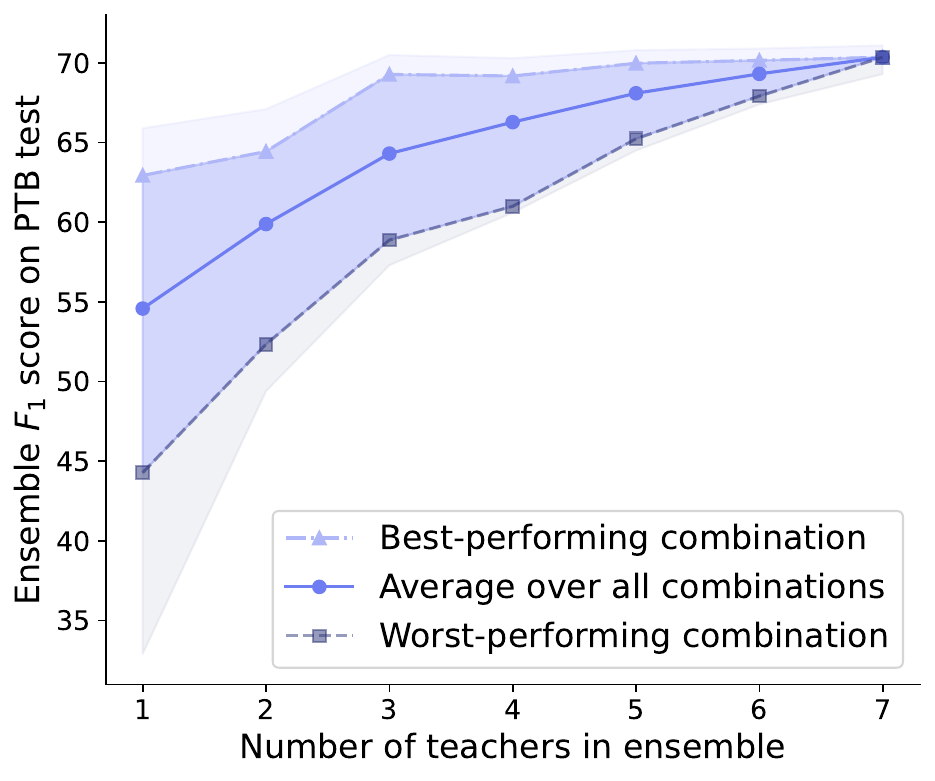}\vspace{-20pt}
    \caption{Ensemble performance with different numbers of teachers. The lines are best-performing, average, and worst-performing combinations. These results are averaged over five runs available in the experiments conducted for Table~\ref{tab:main_results}. The gray shades are the best and worst runs.}
    \label{fig:teachercount}\vspace{-10pt}
\end{wrapfigure}

The over-smoothing problem can be verified by checking the entropy, $-\sum_x p(x)\log p(x)$, of a model's predicted distribution $p$. In Table~\ref{tab:Entropy}, we report the mean and standard deviation of the entropy\footnote{The entropy of each run is averaged over 2,412 samples. The calculation of entropy is based on the codebase of \citet{kim-etal-2019-unsupervised}, available at \url{https://github.com/harvardnlp/urnng}} across five runs. Results clearly show that the union distillation leads to a very smooth distribution (very high entropy), which also explains its low performance (Table~\ref{tab:main_results}). On the contrary, our ensemble-then-distill approach yields much lower entropy, providing strong evidence of the alleviation of the over-smoothing problem.

\textbf{Analyzing the number of teachers.} In our main experiment (Table~\ref{tab:main_results}), we perform an ensemble of seven popular unsupervised parsers. We would like to analyze the performance of ensemble models with different numbers of teachers,\footnote{We have $2^7-1$ combinations, which are tractable because our CYK algorithm is efficient.} and results are shown in Figure~\ref{fig:teachercount}.

We see a consistent trend that more teachers lead to higher performance. Profoundly, the top dashed line suggests that, even if we start with a strong teacher, adding weaker teachers also improves, or at least does not hurt, the performance. Further, the decrease in the width of gray shades (deviations of best and worst runs) suggests that more teachers also lead to lower variance. Overall, this analysis conclusively shows that, with a growing number of teachers, our ensemble approach not only improves performance, but also makes unsupervised parsing more robust.

\textbf{Additional results.} We present supplementary analyses in the appendix. \ref{app:efficiency}: Inference efficiency; \ref{app:length}:~Performance by sentence lengths; and \ref{app:breakdown}:~Performance by different constituency types.

\section{Related Work}
\textbf{Unsupervised syntactic structure discovery} carries a long history and has attracted much attention in different ages~\citep{klein2005unsupervised,shen2018ordered,li-lu-2023-contextual}. Its significance lies in the potential to help low-resource domains~\citep{kann-etal-2019-neural} and its important role in cognitive science, such as understanding how children learn language~\citep{Exemplar2Grammar}.
\citet{6137337} show unsupervised constituency parsing methods are not limited to linguistics but can also be used to parse motion-sensor data by treating it as a language. This approach finds an abstraction of motion data and leads to a better understanding of the signal semantics.

Unsupervised syntactic structure discovery can be divided into different tasks: unsupervised constituency parsing, which  organizes the phrases of a sentence in a hierarchical manner~\citep{constituencyTree}; unsupervised dependency parsing~\citep{nivre2010dependency, naseem2010using, han-etal-2020-survey}, which determines the syntactic relation between the words in a sentence; and 
unsupervised chunking, which aims at segmenting a text into groups of syntactically related words in a flattened structure~\citep{deshmukh-etal-2021-unsupervised-chunking, wu2023unsupervised}.

Our work falls in the category of unsupervised constituency parsing. Previous work has proposed various heuristics and indirect supervisions to tackle this task \citep{snyder2009unsupervised,kim-etal-2019-compound,drozdov-etal-2019-unsupervised-latent,shi-etal-2019-visually}, as mentioned in Section~\ref{sec:introduction}.
In our work, we propose to build an ensemble model to utilize the expertise of different unsupervised parsers.

\textbf{Minimum Bayes risk (MBR) decoding} minimizes a Bayes risk (i.e., expected loss) during inference~\citep{MBRbook}. For example, machine translation systems may generate a set of candidate outputs, and define the risk as the dissimilarity between one candidate output and the rest; MBR decoding selects the lowest-risk candidate translation that is most similar to others~\citep{kumar-byrne-2004-minimum,10.1162/tacl_a_00491}. Similar approaches are applied to other decoding tasks, such as speech recognition~\citep{inproceedingsspeechrecognition}, text summarization~\citep{suzgun-etal-2023-follow}, text-to-code translation~\citep{shi-etal-2022-natural}, and dependency parsing~\citep{smith-smith-2007-probabilistic}. For constituency parsing, \citet{titov-henderson-2006-loss} formulate the task under the MBR framework, and \citet{petrov-klein-2007-improved} extend it to state-split PCFGs.

In this work, we develop a novel generative MBR method for ensemble constituency parsing that searches the entire binary tree space by an efficient CYK-like dynamic programming, significantly differing from common MBR approaches that perform selection on a candidate set.

\textbf{Knowledge distillation (KD)} is commonly used to train a small student model from a large teacher model~\citep{sun-etal-2019-patient,jiao-etal-2020-tinybert}. Evidence show that the teacher's predicted probability contains more knowledge than a groundtruth label and can better train the student model~\citep{hinton2015distilling,wen-etal-2023-f}.

Interestingly, KD is originally proposed to train a small model from an ensemble of teachers~\citep{KDoriginal,hinton2015distilling}. They address simple classification tasks and use either voting or average ensembles to train the student. A voting ensemble is similar to MBR, but only works for classification tasks; it cannot be applied to structure prediction (e.g., sequences or trees). An average ensemble takes the average of probabilities; thus, it resembles union distillation, which is the predominant approach for multi-teacher distillation in recent years~\citep{wu-etal-2021-one,yang2020model}.
However, these approaches may suffer from the over-smoothing problem when teachers are heterogeneous (Section~\ref{discussion:uniondistill}).
In our work, we propose a novel MBR-based ensemble method for multi-teacher distillation, which largely alleviates the over-smoothing problem and is able to utilize different teachers' expertise.

\section{Conclusion}
\vspace{-.3cm}

In this work, we reveal an interesting phenomenon that different unsupervised parsers learn different expertise, and we propose a novel ensemble approach by introducing a new notion of ``tree averaging'' to leverage such heterogeneous expertise. Further, we distill the ensemble knowledge into a student model to improve inference efficiency; the proposed ensemble-then-distill approach also addresses the over-smoothing problem in multi-teacher distillation. Overall, our method shows consistent effectiveness with various teacher models and is robust in the domain-shift setting, largely bridging the gap between supervised and unsupervised constituency parsing. We will discuss future work in Appendix~\ref{app:future}.

\bibliography{iclr2024_conference}
\bibliographystyle{iclr2024_conference}

\newpage

\appendix
\section{Our CYK Variant}
\label{apdx:CYKalgo}

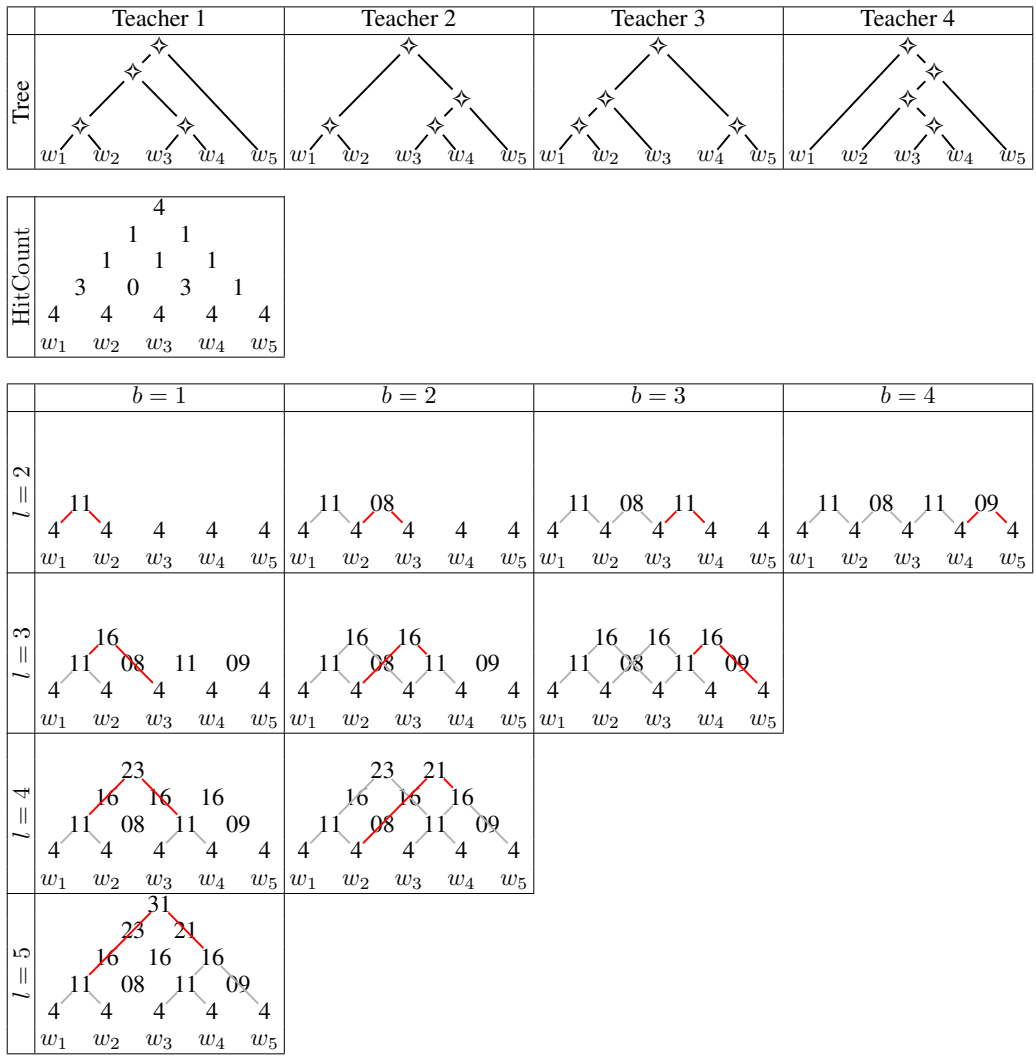
\begin{figure}[!b]
\resizebox{\linewidth}{!}{
\renewcommand{\tabcolsep}{2pt}
    \centering
    \begin{tabular}{|c|c@{}c@{}c@{}c@{}c@{}c@{}c@{}c@{}c|c@{}c@{}c@{}c@{}c@{}c@{}c@{}c@{}c|c@{}c@{}c@{}c@{}c@{}c@{}c@{}c@{}c|c@{}c@{}c@{}c@{}c@{}c@{}c@{}c@{}c|}
    \hline
    & \multicolumn{9}{c|}{Teacher 1} & \multicolumn{9}{c|}{Teacher 2} & \multicolumn{9}{c|}{Teacher 3} & \multicolumn{9}{c|}{Teacher 4}\\
    \hline
    \multirow{5}{*}{\rotatebox[origin=c]{90}{Tree}}
    & &&&&\tikzmarknode{t1w12345}{\ding{71}}&&&&
    & &&&&\tikzmarknode{t2w12345}{\ding{71}}&&&&
    & &&&&\tikzmarknode{t3w12345}{\ding{71}}&&&&
    & &&&&\tikzmarknode{t4w12345}{\ding{71}}&&&&
    \\
    & &&&\tikzmarknode{t1w1234}{\ding{71}}&&\widthof{\hphantom{\ding{71}}}&&&
    & &&&\tikzmarknode{t2w1234}{\hphantom{\ding{71}}}&&\tikzmarknode{t2w2345}{\hphantom{\ding{71}}}&&&
    & &&&\tikzmarknode{t3w1234}{\hphantom{\ding{71}}}&&\tikzmarknode{t3w2345}{\hphantom{\ding{71}}}&&&
    & &&&\tikzmarknode{t4w1234}{\hphantom{\ding{71}}}&&\tikzmarknode{t4w2345}{\ding{71}}&&&
    \\
    & &&\tikzmarknode{t1w123}{\hphantom{\ding{71}}}&&\tikzmarknode{t1w234}{\hphantom{\ding{71}}}&&\tikzmarknode{t1w345}{\hphantom{\ding{71}}}&&
    & &&\tikzmarknode{t2w123}{\hphantom{\ding{71}}}&&\tikzmarknode{t2w234}{\hphantom{\ding{71}}}&&\tikzmarknode{t2w345}{\ding{71}}&&
    & &&\tikzmarknode{t3w123}{\ding{71}}&&\tikzmarknode{t3w234}{\hphantom{\ding{71}}}&&\tikzmarknode{t3w345}{\hphantom{\ding{71}}}&&
    & &&\tikzmarknode{t4w123}{\hphantom{\ding{71}}}&&\tikzmarknode{t4w234}{\ding{71}}&&\tikzmarknode{t4w345}{\hphantom{\ding{71}}}&&
    \\
    & &\tikzmarknode{t1w12}{\ding{71}}&&\tikzmarknode{t1w23}{\hphantom{\ding{71}}}&&\tikzmarknode{t1w34}{\ding{71}}&&\tikzmarknode{t1w45}{\hphantom{\ding{71}}}&
    & &\tikzmarknode{t2w12}{\ding{71}}&&\tikzmarknode{t2w23}{\hphantom{\ding{71}}}&&\tikzmarknode{t2w34}{\ding{71}}&&\tikzmarknode{t2w45}{\hphantom{\ding{71}}}&
    & &\tikzmarknode{t3w12}{\ding{71}}&&\tikzmarknode{t3w23}{\hphantom{\ding{71}}}&&\tikzmarknode{t3w34}{\hphantom{\ding{71}}}&&\tikzmarknode{t3w45}{\ding{71}}&
    & &\tikzmarknode{t4w12}{\hphantom{\ding{71}}}&&\tikzmarknode{t4w23}{\hphantom{\ding{71}}}&&\tikzmarknode{t4w34}{\ding{71}}&&\tikzmarknode{t4w45}{\hphantom{\ding{71}}}&
    \\
    &
    \tikzmarknode{t1w1}{$w_1$}&&\tikzmarknode{t1w2}{$w_2$}&&\tikzmarknode{t1w3}{$w_3$}&&\tikzmarknode{t1w4}{$w_4$}&&\tikzmarknode{t1w5}{$w_5$} & 
    \tikzmarknode{t2w1}{$w_1$}&&\tikzmarknode{t2w2}{$w_2$}&&\tikzmarknode{t2w3}{$w_3$}&&\tikzmarknode{t2w4}{$w_4$}&&\tikzmarknode{t2w5}{$w_5$} & 
    \tikzmarknode{t3w1}{$w_1$}&&\tikzmarknode{t3w2}{$w_2$}&&\tikzmarknode{t3w3}{$w_3$}&&\tikzmarknode{t3w4}{$w_4$}&&\tikzmarknode{t3w5}{$w_5$} & 
    \tikzmarknode{t4w1}{$w_1$}&&\tikzmarknode{t4w2}{$w_2$}&&\tikzmarknode{t4w3}{$w_3$}&&\tikzmarknode{t4w4}{$w_4$}&&\tikzmarknode{t4w5}{$w_5$}
    \\\hline
    \multicolumn{0}{c}{}
    \\\cline{0-9}
    \multirow{6}{*}{\rotatebox[origin=c]{90}{$\operatorname{HitCount}$}}
    & &&&&4&&&&\\
    & &&&1&&1&&&\\
    & &&1&&1&&1&&\\
    & &3&&0&&3&&1&\\
    & 4&&4&&4&&4&&4\\
    &
    {$w_1$}&&{$w_2$}&&{$w_3$}&&{$w_4$}&&{$w_5$}
    \\\cline{0-9}
    \multicolumn{0}{c}{}
    \\\hline
    & \multicolumn{9}{c|}{$b=1$} & \multicolumn{9}{c|}{$b=2$} & \multicolumn{9}{c|}{$b=3$} & \multicolumn{9}{c|}{$b=4$} \\
    \hline
    \multirow{6}{*}{\rotatebox[origin=c]{90}{$l=2$}}
    & &&&&&&&& & &&&&&&&& & &&&&&&&& & &&&&&&&&\\
    & &&&&&&&& & &&&&&&&& & &&&&&&&& & &&&&&&&&\\
    & &&&&&&&& & &&&&&&&& & &&&&&&&& & &&&&&&&&\\
    & &\tikzmarknode{l2b1w12}{11}&&&&&&& & &\tikzmarknode{l2b2w12}{11}&&\tikzmarknode{l2b2w23}{08}&&&&& & &\tikzmarknode{l2b3w12}{11}&&\tikzmarknode{l2b3w23}{08}&&\tikzmarknode{l2b3w34}{11}&&& & &\tikzmarknode{l2b4w12}{11}&&\tikzmarknode{l2b4w23}{08}&&\tikzmarknode{l2b4w34}{11}&&\tikzmarknode{l2b4w45}{09}&\\
    & \tikzmarknode{l2b1w1}{4}&&\tikzmarknode{l2b1w2}{4}&&\tikzmarknode{l2b1w3}{4}&&\tikzmarknode{l2b1w4}{4}&&\tikzmarknode{l2b1w5}{4} & \tikzmarknode{l2b2w1}{4}&&\tikzmarknode{l2b2w2}{4}&&\tikzmarknode{l2b2w3}{4}&&\tikzmarknode{l2b2w4}{4}&&\tikzmarknode{l2b2w5}{4} & \tikzmarknode{l2b3w1}{4}&&\tikzmarknode{l2b3w2}{4}&&\tikzmarknode{l2b3w3}{4}&&\tikzmarknode{l2b3w4}{4}&&\tikzmarknode{l2b3w5}{4} & \tikzmarknode{l2b4w1}{4}&&\tikzmarknode{l2b4w2}{4}&&\tikzmarknode{l2b4w3}{4}&&\tikzmarknode{l2b4w4}{4}&&\tikzmarknode{l2b4w5}{4}\\
    &
    {$w_1$}&&{$w_2$}&&{$w_3$}&&{$w_4$}&&{$w_5$} & 
    {$w_1$}&&{$w_2$}&&{$w_3$}&&{$w_4$}&&{$w_5$} & 
    {$w_1$}&&{$w_2$}&&{$w_3$}&&{$w_4$}&&{$w_5$} & 
    {$w_1$}&&{$w_2$}&&{$w_3$}&&{$w_4$}&&{$w_5$}\\
    \cline{0-36}
    \multirow{6}{*}{\rotatebox[origin=c]{90}{$l=3$}}
    & &&&&&&&& & &&&&&&&& & &&&&&&&& & \\
    & &&&&&&&& & &&&&&&&& & &&&&&&&& & \\
    & &&\tikzmarknode{l3b1w123}{16}&&&&&& & &&\tikzmarknode{l3b2w123}{16}&&\tikzmarknode{l3b2w234}{16}&&&& & &&\tikzmarknode{l3b3w123}{16}&&\tikzmarknode{l3b3w234}{16}&&\tikzmarknode{l3b3w345}{16}&& & \\
    & &\tikzmarknode{l3b1w12}{11}&&\tikzmarknode{l3b1w23}{08}&&\tikzmarknode{l3b1w34}{11}&&\tikzmarknode{l3b1w45}{09}& & &\tikzmarknode{l3b2w12}{11}&&\tikzmarknode{l3b2w23}{08}&&\tikzmarknode{l3b2w34}{11}&&\tikzmarknode{l3b2w45}{09}& & &\tikzmarknode{l3b3w12}{11}&&\tikzmarknode{l3b3w23}{08}&&\tikzmarknode{l3b3w34}{11}&&\tikzmarknode{l3b3w45}{09}& & \\
    & \tikzmarknode{l3b1w1}{4}&&\tikzmarknode{l3b1w2}{4}&&\tikzmarknode{l3b1w3}{4}&&\tikzmarknode{l3b1w4}{4}&&\tikzmarknode{l3b1w5}{4} & \tikzmarknode{l3b2w1}{4}&&\tikzmarknode{l3b2w2}{4}&&\tikzmarknode{l3b2w3}{4}&&\tikzmarknode{l3b2w4}{4}&&\tikzmarknode{l3b2w5}{4} & \tikzmarknode{l3b3w1}{4}&&\tikzmarknode{l3b3w2}{4}&&\tikzmarknode{l3b3w3}{4}&&\tikzmarknode{l3b3w4}{4}&&\tikzmarknode{l3b3w5}{4} & \\
    &
    {$w_1$}&&{$w_2$}&&{$w_3$}&&{$w_4$}&&{$w_5$} & 
    {$w_1$}&&{$w_2$}&&{$w_3$}&&{$w_4$}&&{$w_5$} & 
    {$w_1$}&&{$w_2$}&&{$w_3$}&&{$w_4$}&&{$w_5$}\\
    \cline{0-27}
    \multirow{6}{*}{\rotatebox[origin=c]{90}{$l=4$}}
    & &&&&&&&& & &&&&&&&& & \\
    & &&&\tikzmarknode{l4b1w1234}{23}&&&&& & &&&\tikzmarknode{l4b2w1234}{23}&&\tikzmarknode{l4b2w2345}{21}&&& & \\
    & &&\tikzmarknode{l4b1w123}{16}&&\tikzmarknode{l4b1w234}{16}&&\tikzmarknode{l4b1w345}{16}&& & &&\tikzmarknode{l4b2w123}{16}&&\tikzmarknode{l4b2w234}{16}&&\tikzmarknode{l4b2w345}{16}&& & \\
    & &\tikzmarknode{l4b1w12}{11}&&\tikzmarknode{l4b1w23}{08}&&\tikzmarknode{l4b1w34}{11}&&\tikzmarknode{l4b1w45}{09}& & &\tikzmarknode{l4b2w12}{11}&&\tikzmarknode{l4b2w23}{08}&&\tikzmarknode{l4b2w34}{11}&&\tikzmarknode{l4b2w45}{09}& & \\
    & \tikzmarknode{l4b1w1}{4}&&\tikzmarknode{l4b1w2}{4}&&\tikzmarknode{l4b1w3}{4}&&\tikzmarknode{l4b1w4}{4}&&\tikzmarknode{l4b1w5}{4} & \tikzmarknode{l4b2w1}{4}&&\tikzmarknode{l4b2w2}{4}&&\tikzmarknode{l4b2w3}{4}&&\tikzmarknode{l4b2w4}{4}&&\tikzmarknode{l4b2w5}{4} \\
    &
    {$w_1$}&&{$w_2$}&&{$w_3$}&&{$w_4$}&&{$w_5$} & 
    {$w_1$}&&{$w_2$}&&{$w_3$}&&{$w_4$}&&{$w_5$}\\
    \cline{0-18}
    \multirow{6}{*}{\rotatebox[origin=c]{90}{$l=5$}}
    & &&&&\tikzmarknode{l5b1w12345}{31}&&&& \\
    & &&&\tikzmarknode{l5b1w1234}{23}&&\tikzmarknode{l5b1w2345}{21}&&& \\
    & &&\tikzmarknode{l5b1w123}{16}&&\tikzmarknode{l5b1w234}{16}&&\tikzmarknode{l5b1w345}{16}&& \\
    & &\tikzmarknode{l5b1w12}{11}&&\tikzmarknode{l5b1w23}{08}&&\tikzmarknode{l5b1w34}{11}&&\tikzmarknode{l5b1w45}{09}& \\
    & \tikzmarknode{l5b1w1}{4}&&\tikzmarknode{l5b1w2}{4}&&\tikzmarknode{l5b1w3}{4}&&\tikzmarknode{l5b1w4}{4}&&\tikzmarknode{l5b1w5}{4} \\
    &
    {$w_1$}&&{$w_2$}&&{$w_3$}&&{$w_4$}&&{$w_5$}\\
    \cline{0-9}
\end{tabular}
    \begin{tikzpicture}[remember picture, overlay]
        \draw[-, black, thick] (t1w12345) -- (t1w1234);
        \draw[-, black, thick] (t1w12345) -- (t1w5);
        \draw[-, black, thick] (t1w1234) -- (t1w12);
        \draw[-, black, thick] (t1w1234) -- (t1w34);
        \draw[-, black, thick] (t1w12) -- (t1w1);
        \draw[-, black, thick] (t1w12) -- (t1w2);
        \draw[-, black, thick] (t1w34) -- (t1w3);
        \draw[-, black, thick] (t1w34) -- (t1w4);
        
        \draw[-, black, thick] (t2w12345) -- (t2w12);
        \draw[-, black, thick] (t2w12345) -- (t2w345);
        \draw[-, black, thick] (t2w12) -- (t2w1);
        \draw[-, black, thick] (t2w12) -- (t2w2);
        \draw[-, black, thick] (t2w345) -- (t2w34);
        \draw[-, black, thick] (t2w345) -- (t2w5);
        \draw[-, black, thick] (t2w34) -- (t2w3);
        \draw[-, black, thick] (t2w34) -- (t2w4);
        
        \draw[-, black, thick] (t3w12345) -- (t3w123);
        \draw[-, black, thick] (t3w12345) -- (t3w45);
        \draw[-, black, thick] (t3w123) -- (t3w12);
        \draw[-, black, thick] (t3w123) -- (t3w3);
        \draw[-, black, thick] (t3w12) -- (t3w1);
        \draw[-, black, thick] (t3w12) -- (t3w2);
        \draw[-, black, thick] (t3w45) -- (t3w4);
        \draw[-, black, thick] (t3w45) -- (t3w5);
        
        \draw[-, black, thick] (t4w12345) -- (t4w1);
        \draw[-, black, thick] (t4w12345) -- (t4w2345);
        \draw[-, black, thick] (t4w2345) -- (t4w234);
        \draw[-, black, thick] (t4w2345) -- (t4w5);
        \draw[-, black, thick] (t4w234) -- (t4w2);
        \draw[-, black, thick] (t4w234) -- (t4w34);
        \draw[-, black, thick] (t4w34) -- (t4w3);
        \draw[-, black, thick] (t4w34) -- (t4w4);
        
        \draw[-, red, thick] (l2b1w12) -- (l2b1w1);
        \draw[-, red, thick] (l2b1w12) -- (l2b1w2);
        
        \draw[-, black!30, thick] (l2b2w12) -- (l2b2w1);
        \draw[-, black!30, thick] (l2b2w12) -- (l2b2w2);
        \draw[-, red, thick] (l2b2w23) -- (l2b2w2);
        \draw[-, red, thick] (l2b2w23) -- (l2b2w3);
        
        \draw[-, black!30, thick] (l2b3w12) -- (l2b3w1);
        \draw[-, black!30, thick] (l2b3w12) -- (l2b3w2);
        \draw[-, black!30, thick] (l2b3w23) -- (l2b3w2);
        \draw[-, black!30, thick] (l2b3w23) -- (l2b3w3);
        \draw[-, red, thick] (l2b3w34) -- (l2b3w3);
        \draw[-, red, thick] (l2b3w34) -- (l2b3w4);
        
        \draw[-, black!30, thick] (l2b4w12) -- (l2b4w1);
        \draw[-, black!30, thick] (l2b4w12) -- (l2b4w2);
        \draw[-, black!30, thick] (l2b4w23) -- (l2b4w2);
        \draw[-, black!30, thick] (l2b4w23) -- (l2b4w3);
        \draw[-, black!30, thick] (l2b4w34) -- (l2b4w3);
        \draw[-, black!30, thick] (l2b4w34) -- (l2b4w4);
        \draw[-, red, thick] (l2b4w45) -- (l2b4w4);
        \draw[-, red, thick] (l2b4w45) -- (l2b4w5);
        
        \draw[-, black!30, thick] (l3b1w12) -- (l3b1w1);
        \draw[-, black!30, thick] (l3b1w12) -- (l3b1w2);
        \draw[-, red, thick] (l3b1w123) -- (l3b1w12);
        \draw[-, red, thick] (l3b1w123) -- (l3b1w3);
        
        \draw[-, black!30, thick] (l3b2w12) -- (l3b2w1);
        \draw[-, black!30, thick] (l3b2w12) -- (l3b2w2);
        \draw[-, black!30, thick] (l3b2w123) -- (l3b2w12);
        \draw[-, black!30, thick] (l3b2w123) -- (l3b2w3);
        \draw[-, black!30, thick] (l3b2w34) -- (l3b2w3);
        \draw[-, black!30, thick] (l3b2w34) -- (l3b2w4);
        \draw[-, red, thick] (l3b2w234) -- (l3b2w2);
        \draw[-, red, thick] (l3b2w234) -- (l3b2w34);
        
        \draw[-, black!30, thick] (l3b3w12) -- (l3b3w1);
        \draw[-, black!30, thick] (l3b3w12) -- (l3b3w2);
        \draw[-, black!30, thick] (l3b3w123) -- (l3b3w12);
        \draw[-, black!30, thick] (l3b3w123) -- (l3b3w3);
        \draw[-, black!30, thick] (l3b3w34) -- (l3b3w3);
        \draw[-, black!30, thick] (l3b3w34) -- (l3b3w4);
        \draw[-, black!30, thick] (l3b3w234) -- (l3b3w2);
        \draw[-, black!30, thick] (l3b3w234) -- (l3b3w34);
        \draw[-, red, thick] (l3b3w345) -- (l3b3w34);
        \draw[-, red, thick] (l3b3w345) -- (l3b3w5);

        \draw[-, black!30, thick] (l4b1w12) -- (l4b1w1);
        \draw[-, black!30, thick] (l4b1w12) -- (l4b1w2);
        \draw[-, black!30, thick] (l4b1w34) -- (l4b1w3);
        \draw[-, black!30, thick] (l4b1w34) -- (l4b1w4);
        \draw[-, red, thick] (l4b1w1234) -- (l4b1w12);
        \draw[-, red, thick] (l4b1w1234) -- (l4b1w34);

        \draw[-, black!30, thick] (l4b2w12) -- (l4b2w1);
        \draw[-, black!30, thick] (l4b2w12) -- (l4b2w2);
        \draw[-, black!30, thick] (l4b2w34) -- (l4b2w3);
        \draw[-, black!30, thick] (l4b2w34) -- (l4b2w4);
        \draw[-, black!30, thick] (l4b2w1234) -- (l4b2w12);
        \draw[-, black!30, thick] (l4b2w1234) -- (l4b2w34);
        \draw[-, black!30, thick] (l4b2w345) -- (l4b2w34);
        \draw[-, black!30, thick] (l4b2w345) -- (l4b2w5);
        \draw[-, red, thick] (l4b2w2345) -- (l4b2w2);
        \draw[-, red, thick] (l4b2w2345) -- (l4b2w345);

        \draw[-, black!30, thick] (l5b1w345) -- (l5b1w34);
        \draw[-, black!30, thick] (l5b1w345) -- (l5b1w5);
        \draw[-, black!30, thick] (l5b1w34) -- (l5b1w3);
        \draw[-, black!30, thick] (l5b1w34) -- (l5b1w4);
        \draw[-, black!30, thick] (l5b1w12) -- (l5b1w1);
        \draw[-, black!30, thick] (l5b1w12) -- (l5b1w2);
        \draw[-, red, thick] (l5b1w12345) -- (l5b1w12);
        \draw[-, red, thick] (l5b1w12345) -- (l5b1w345);
    \end{tikzpicture}
}
    \caption{Step-by-step illustration of our CYK algorithm, showing the dynamic changes in the $H$ along with the construction of the corresponding optimal binary constituency tree.}
    \label{fig:illustration}
\end{figure}
In this appendix, we provide a step-by-step illustration of our CYK-based ensemble algorithm introduced in Section~\ref{Sec:OurCYK}.

Consider four teachers predicting the trees in the first row of Figure~\ref{fig:illustration}. The hit count of each span is shown in the second row. For example, the span $(w_1w_2)$ hits $3$ times, namely, Teachers 1--3. 

The initialization of the algorithm is to obtain the total hit count for a single word, which is simply the same as the number of teachers because every word appears intact in every teacher's prediction. The initialization has five cells in a row, and is omitted in the figure to fit the page width.

For recursion, we first consider the constituents of two words, denoted by $l=2$. A constituent's total hit count, denoted by $H_{b:e}$ in Eqn.~(\ref{eq:H}), inherits those of its children, plus its own hit count. In the cell of $l=2,b=1$, for example, $H_{1:3}=4+4+3=11$, where $3$ is the hit count of the span $(w_1w_2)$, shown before. 

For the next step of recursion, we consider three-word constituents, i.e., $l=3$. For example, the span $w_1 w_2 w_3$ has two possible tree structures $(w_1(w_2w_3))$ and $((w_1w_2)w_3)$. The former leads to a total hit count of $13$, whereas the latter leads to $16$. Therefore, $((w_1w_2)w_3)$ is chosen, with the best total hit count $H_{1:4}=16$.

The process is repeated until we have the best parse tree of the whole sentence, which is $l=5$ for the 5-word sentence in Figure~\ref{fig:illustration}.

We provide the pseudocode for the process in Algorithm~\ref{algo:CYK}.

\begin{algorithm}[!t]
\caption{Our CYK Variant}
\label{algo:CYK}
\begin{algorithmic}[1]
    \item \textbf{input:} $s, \{T_i\}_{i=1}^K$
    \For{$b \gets 1$ to $|s|$} $\qquad \rhd$ Base cases
        \State $H_{b:b+1} = K$
        \State $L_{b:b+1}=\{s_{b:b+1}\}$
    \EndFor
    \For{$l \gets 2$ to $|s|$} $\qquad\hspace{0.06cm}\rhd$Iterate over different lengths of constituents
        \For{$b \gets 1$ to $|s|-l+1$} $\qquad \rhd$ Iterate over different possible constituents of length $l$
            \State $e \gets b+l$
            \State $j^*_{s:b} \gets \argmax\limits_{b < j < e} ( H_{b:j}+H_{j:e} {\color{gray}  + \operatorname{HitCount}(s_{b:e}, \{T_i(s)\}_{i=1}^K)})$
        \Statex \hspace{4cm}$\rhd$ The {\color{gray}gray} term need not be implemented as it is a constant in $j$
            \State $H_{b:e} \gets H_{b:j^*_{s:b}} + H_{j^*_{s:b}:e} + \operatorname{HitCount}(s_{b:e}, \{T_i(s)\}_{i=1}^K)$
            \State $L_{b:e} \gets L_{b:j^*_{s:b}} \cup L_{j^*_{s:b}:e} \cup \{s_{b:e}\}$
        \EndFor
    \EndFor
    \State \Return $L_{1:|s|+1}$
\end{algorithmic}
\end{algorithm}

\section{Supplementary Analyses}
\label{app:additional}

\subsection{Inference Efficiency}\label{app:efficiency}

\begin{wraptable}{r}{0.45\textwidth}
    \centering
\resizebox{\linewidth}{!}{
    \begin{tabular}{l@{\qquad} l r r}
        \toprule
        &&\multicolumn{2}{c}{Inference Time (ms)}\\
        & Model & w/ GPU & w/o GPU \\
        \midrule
        \multicolumn{4}{l}{Teachers}\\
        & ON & 35 & 130 \\
        & Neural PCFG & 610 & 630 \\
        & Compound PCFG & 560 & 590 \\
        & DIORA & 30 & 30 \\
        & S-DIORA & 110 & 140 \\
        & ConTest & 4,300 & 59,500 \\
        & ContexDestort & 1,890 & 11,110 \\
        \multicolumn{4}{l}{Our ensemble} \\
        & CYK part & 6 & 6 \\
        & Total & 7,541 & 72,136 \\
        \multicolumn{4}{l}{Student} \\
        & RNNG & 410 & 410 \\
        \bottomrule
    \end{tabular}
}\vspace{-5pt}
    \caption{Per-sample inference time (in milliseconds) on the PTB test. }
    \label{tab:time_analysis}
\end{wraptable}

We propose to distill the ensemble knowledge into a student model to increase the inference efficiency. We conducted an analysis on the inference time of different approaches, where we measured the run time using 28 Intel(R) Core(TM) i9-9940X (@3.30GHz) CPUs with or without GPU (Nvidia RTX Titan). Table~\ref{tab:time_analysis} reports the average time elapsed for performing inference on one sample\footnote{The average time was computed on 100 samples of the PTB test set, due to the slow inference of certain teachers without GPU.} of the PTB test set, ignoring loading models, reading inputs, and writing outputs. 

In the table, the total inference time of our ensemble model is the summation of all the teachers and the CYK algorithm. As expected, an ensemble approach is slow because it has to perform inference for every teacher. However, our CYK-based ensemble algorithm is extremely efficient and its inference time is negligible compared with the teacher models.

The RNNG student model learns the knowledge from the cumbersome ensemble, and is able to perform inference efficiently with an 18x and 175x speedup with and without GPU, respectively. This shows the necessity of having knowledge distillation on top of the ensemble. Overall, RNNG achieves comparable performance to its ensemble teacher (Tables~\ref{tab:main_results} and~\ref{tab:susanne_results}) but drastically speeds up the inference, being a useful model in practice.

\newpage

\subsection{Performance by Sentence Lengths}\label{app:length}

\begin{figure}[t]
    \centering\includegraphics[width=.65\linewidth]{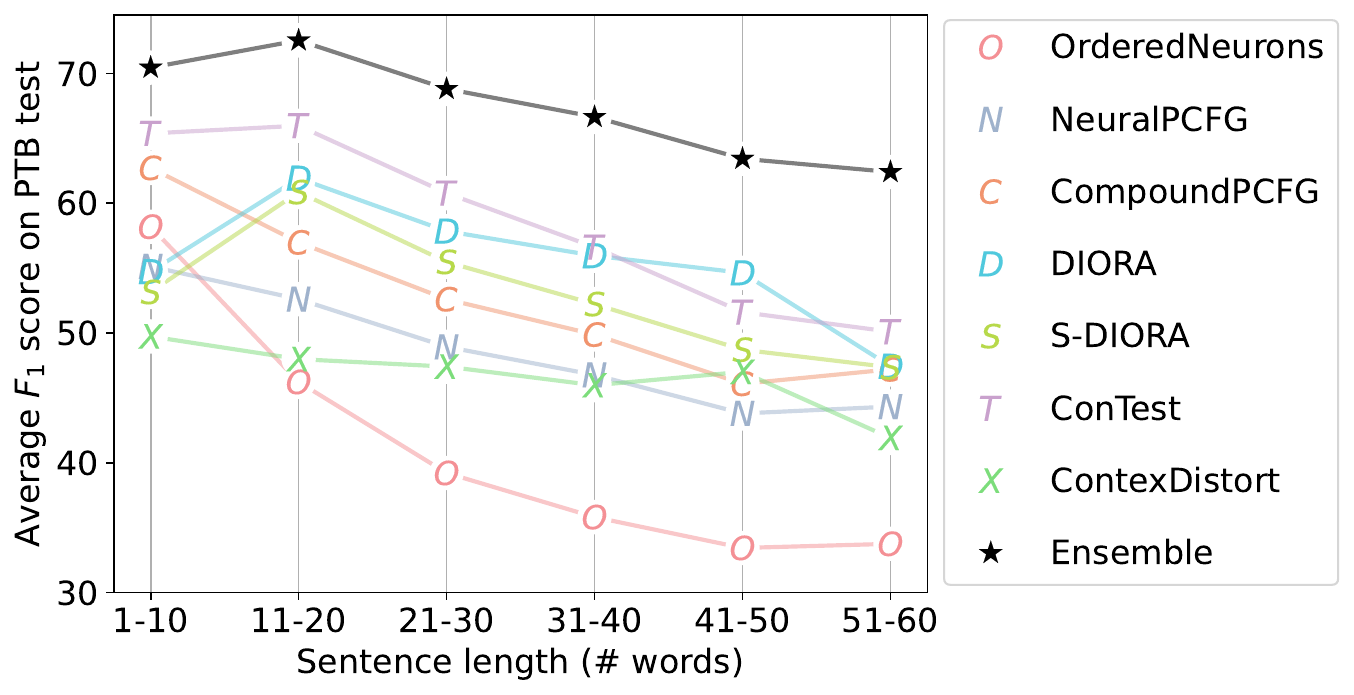}\vspace{-7pt}
    \caption{Performance by sentence lengths. $F_1$ scores are averaged over five different runs.}
    \label{fig:perlength}
\end{figure}

Figure~\ref{fig:perlength} illustrates the parsing performance on sentences of varying lengths. The result shows that existing unsupervised parsers have different levels of susceptibility to long sentences. For example, ContexDistort shows notable robustness to the length, whereas the performance of Ordered Neurons drops significantly when the sentences are longer. Our ensemble method achieves both high performance and robustness across different lengths.

\subsection{Performance by Constituency Labels}\label{app:breakdown}

\begin{figure}[!b]
\begin{center}
\includegraphics[width=\linewidth]{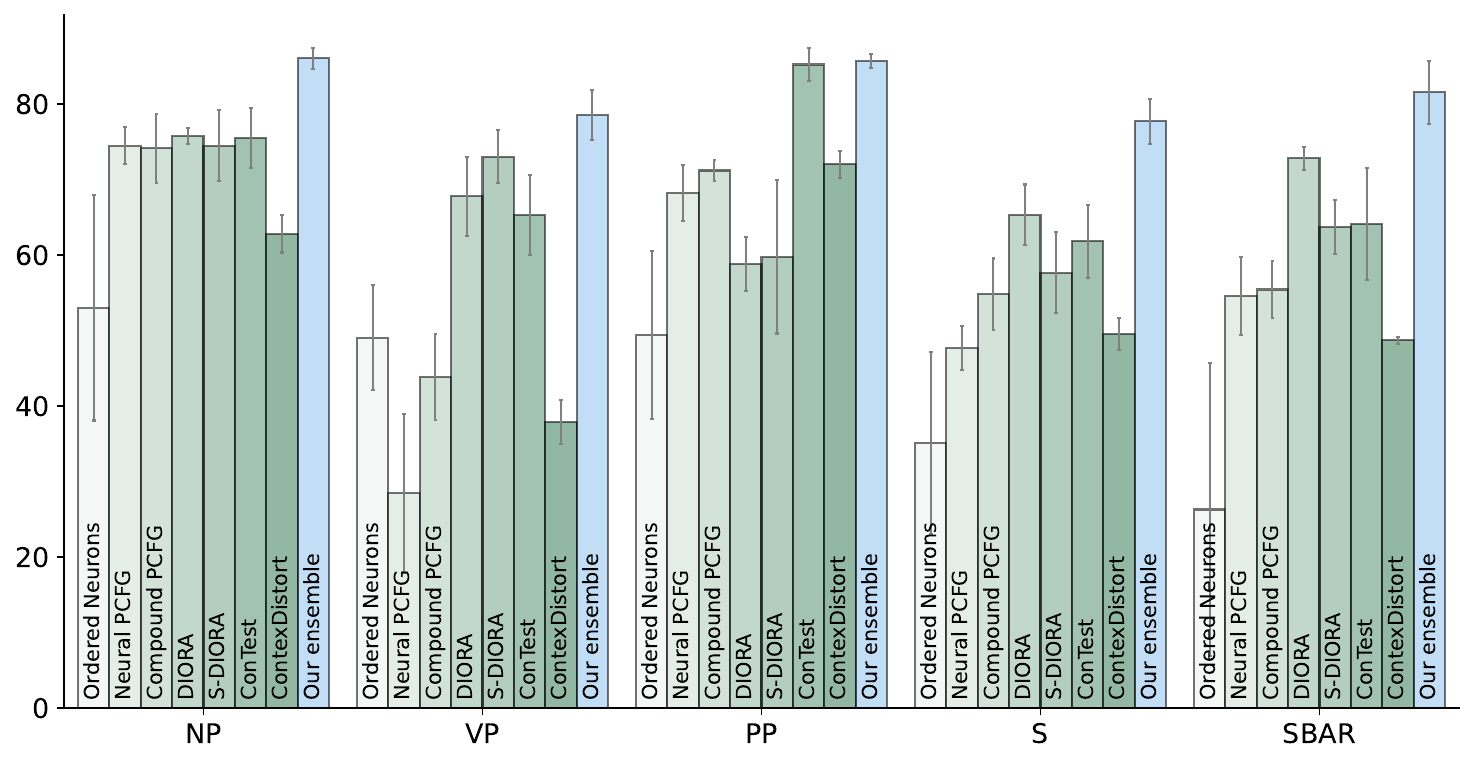}
\end{center}\vspace{-10pt}
\caption{Performance by constituency labels on the PTB test set. Results are measured by recall, because the predicted parse trees are unlabeled; thus, precision and $F_1$ scores cannot be computed~\citep{drozdov-etal-2019-unsupervised-latent}. Bars and gray intervals are the mean and standard deviation, respectively, over five runs.}
\label{fig:pertag}
\end{figure}

In this work, we see different unsupervised parsers learn different patterns (Table~\ref{tab:correlation-table}), and their expertise can be utilized by an ensemble approach (Section~\ref{ss:analysis}). From the linguistic point of view, we are curious about whether there is a relation between such different expertise and the linguistic constituent labels (e.g., noun phrases and verb phrases). 

With this motivation, we report in Figure~\ref{fig:pertag} the breakdown performance by constituency labels, where the most common five labels---namely, noun phrases, propositional phrases, verb phrases, simple declarative clauses, and subordinating conjunction clauses---are considered, covering 95\% of the cases in the PTB test set. Notice that the predicted constituency parse trees are still unlabeled (without tags like noun phrases), whereas the groundtruth constituency labels are used for collecting the statistics. Consequently, only recall scores can be calculated in per-label performance analysis~\citep{drozdov-etal-2019-unsupervised-latent,kim-etal-2019-compound,cao-etal-2020-unsupervised}. 

As seen, existing unsupervised parsers indeed exhibit variations in the performance of different constituency labels. For example, ConTest achieves high performance of prepositional phrases, whereas DIORA works well for clauses (including simple declarative clauses and subordinating conjunction clauses); for noun phrases, most models perform similarly. By contrast, our ensemble model achieves outstanding performance similar to or higher than the best teacher in each category. This provides further evidence that our ensemble model utilizes different teachers' expertise.

\section{Case Studies}

\begin{figure}[t]
\begin{center}
\includegraphics[width=\linewidth]{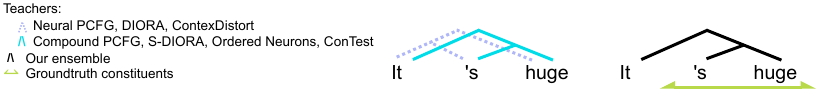}
\end{center}\vspace{-15pt}
\caption{A case study, which shows that voting selects more commonly agreed structures.}
\label{fig:casestudy-local}
\end{figure}

\begin{figure}[t]
\begin{center}
\includegraphics[width=\linewidth]{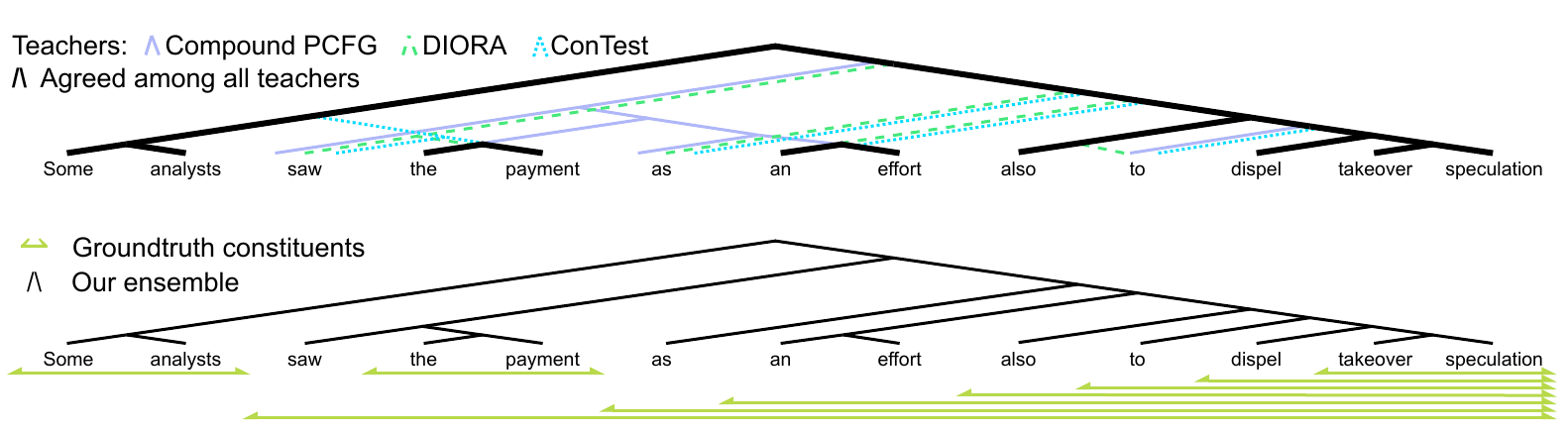}
\end{center}\vspace{-10pt}
\caption{A case study, where the ensemble outperforms all the teachers and achieves 100\% recall over the groundtruth constituents.}\label{fig:casestudy-100recall}
\end{figure}

In this section, we present case studies to show how the ensemble improves the performance. In particular, Figure~\ref{fig:casestudy-local} illustrates teachers' performance, their ensemble output, and groundtruth for the sentence ``It's huge.'' This example represents how voting over local structures may result in correct structure detection. True constituents have a higher chance to appear in the majority of the teachers' outputs. This phenomenon extends to longer sentences and more complex structures. Figure~\ref{fig:casestudy-100recall} presents an example where the ensemble outperforms all its teachers, hitting all the groundtruth constituents, which never happens in any teacher. Note that in this example, every constituent captured by the ensemble appears in at least two out of three teachers.

\begin{figure}[t]
\begin{center}
\includegraphics[width=\linewidth]{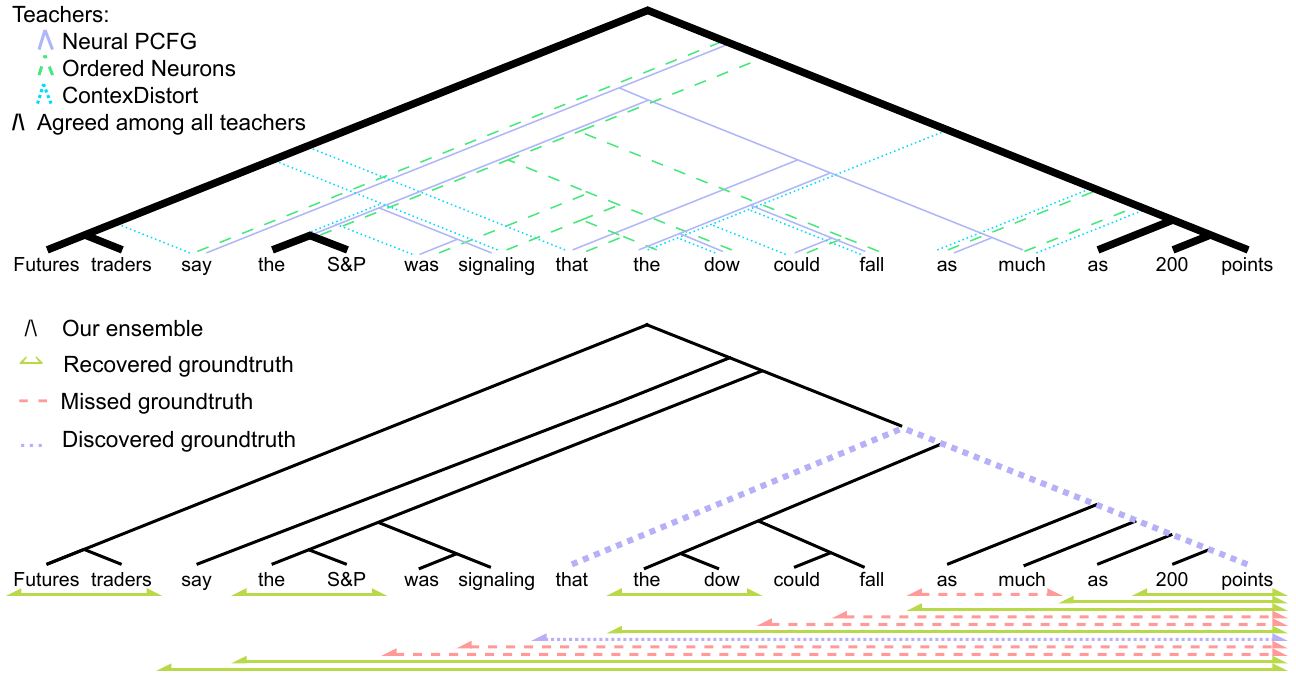}
\end{center}\vspace{-10pt}
\caption{A case study, where the ensemble recovers a true constituent never seen in any teacher's output.}\label{fig:casestudy-discovery}
\end{figure}

Figure~\ref{fig:casestudy-discovery} illustrates a more interesting behavior of the ensemble, where it recovers a true constituent never seen in any teacher's output, drawn in dotted purple in the bottom figure. It happens in complex structures when teachers agree on some local structures but do not agree over the entire sentence. In that case, the ensemble eventually picks the agreed structures and fills the gaps with the remaining options.

\section{Future Work}\label{app:future}
{Future work} may be considered from both linguistic and machine learning perspectives. The proposed ensemble method largely bridges the gap between supervised and unsupervised parsing of the English language. A future direction is to address unsupervised linguistic structure discovery in low-resource and multilingual settings~\citep{shayegh2024disco}. Regarding the machine learning aspect, our work demonstrates the importance of addressing the over-smoothing problem in multi-teacher distillation, and we expect our ensemble-then-distill approach can be extended to different data types, such as sequences and graphs, with proper design of data-specific ensemble methods~\citep{wen2024ebbs}.

\subsubsection*{Acknowledgments}
We would like to thank all reviewers and chairs for their valuable and constructive comments. The research is supported in part by the Natural Sciences and Engineering Research Council of Canada (NSERC), a Mitacs Accelerate project, the Amii Fellow Program, the Canada CIFAR AI Chair Program, an Alberta Innovates Program, and the Digital Research Alliance of Canada (alliancecan.ca). We also thank Yongchang Hao for providing advice on the algorithms.

\newpage

\section{Inventory of Teacher Models}
\label{apdx:ExperimentDetail}

Our experiments involve seven existing unsupervised parsers as teachers, each of which has five runs either based on authors' checkpoints or by our replication using authors' codebases. We show the details in Table~\ref{tab:detailed_variants}, where we also quote the mean $F_1$ scores and, if available, max $F_1$ scores reported in 
 respective papers. Overall, we have achieved similar performance to previous work, which shows the success of our replication and establishes a solid foundation for our ensemble research.

\vspace{5pt}
\begin{table}[h]
\centering
\resizebox{.83\linewidth}{!}{
\begin{tabular}{l c | l | c}
& Run & Source & $F_1$\\
\hline
\toprule
\multirow{6}{*}{\rotatebox[origin=c]{90}{Ordered Neurons}} &
\multicolumn{3}{l}{mean $F_{1}=47.7$, max $F_{1}=49.4$ reported in~\citet{shen2018ordered}} \\
\cmidrule(){2-4}
&1 & Our replication using the original codebase\tablefootnote{\url{https://github.com/yikangshen/Ordered-Neurons}\label{footnote:ONcodebase}} ($\text{seed}=0017$)& $44.8$\\
&2 & Our replication using the original codebase\footnoteagain{footnote:ONcodebase} ($\text{seed}=0031$)& $32.9$\\
&3 & Parsed data available in \citet{kim-etal-2019-compound}\tablefootnote{\url{https://github.com/harvardnlp/compound-pcfg}\label{footnote:compoundcodebase}} & $50.0$\\
&4 & Our replication using the original codebase\footnoteagain{footnote:ONcodebase} ($\text{seed}=7214$)& $47.8$\\
&5 & Our replication using the original codebase\footnoteagain{footnote:ONcodebase} ($\text{seed}=1111$)& $45.9$\\
\midrule
\multirow{6}{*}{\rotatebox[origin=c]{90}{Neural PCFG}} &
\multicolumn{3}{l}{mean $F_{1}=50.8$, max $F_{1}=52.6$ reported in~\citet{kim-etal-2019-compound}} \\
\cmidrule(){2-4}
&1 & Our replication using the original codebase\footnoteagain{footnote:compoundcodebase} ($\text{seed}=3435$)& $48.4$\\
&2 & Parsed data available in the original codebase\footnoteagain{footnote:compoundcodebase} & $52.6$\\
&3 & Our replication using the original codebase\footnoteagain{footnote:compoundcodebase} ($\text{seed}=1234$)& $52.1$\\
&4 & Our replication using the original codebase\footnoteagain{footnote:compoundcodebase} ($\text{seed}=1313$)& $52.3$\\
&5 & Our replication using the original codebase\footnoteagain{footnote:compoundcodebase} ($\text{seed}=5555$)& $49.8$\\
\hline
\multirow{6}{*}{\rotatebox[origin=c]{90}{Compound PCFG}} &
\multicolumn{3}{l}{mean $F_{1}=50.8$,  max $F_{1}=52.6$ reported in~\citet{kim-etal-2019-compound}} \\
\cmidrule(){2-4}
&1 & Parsed data available in the original codebase\footnoteagain{footnote:compoundcodebase} & $60.1$\\
&2 & Our replication using the original codebase\footnoteagain{footnote:compoundcodebase} ($\text{seed}=3435$)& $53.9$\\
&3 & Our replication using the original codebase\footnoteagain{footnote:compoundcodebase} ($\text{seed}=1234$)& $55.4$\\
&4 & Our replication using the original codebase\footnoteagain{footnote:compoundcodebase} ($\text{seed}=0887$)& $53.2$\\
&5 & Our replication using the original codebase\footnoteagain{footnote:compoundcodebase} ($\text{seed}=0778$)& $55.0$\\
\hline
\multirow{6}{*}{\rotatebox[origin=c]{90}{DIORA}} &
\multicolumn{3}{l}{mean $F_{1}=56.8$ reported in~\citet{drozdov-etal-2019-unsupervised-latent}} \\
\cmidrule(){2-4}
&1 & The mlp-softmax checkpoint available on the original codebase\tablefootnote{\url{https://github.com/iesl/diora}\label{footnote:dioracodebase}} & $55.4$\\
&2 & Our replication using the original codebase\footnoteagain{footnote:dioracodebase} ($\text{seed}=0035$)& $59.4$\\
&3 & Our replication using the original codebase\footnoteagain{footnote:dioracodebase} ($\text{seed}=0074$)& $60.3$\\
&4 & Our replication using the original codebase\footnoteagain{footnote:dioracodebase} ($\text{seed}=1313$)& $60.5$\\
&5 & Our replication using the original codebase\footnoteagain{footnote:dioracodebase} ($\text{seed}=5555$)& $58.9$\\
\hline
\multirow{6}{*}{\rotatebox[origin=c]{90}{S-DIORA}} &
\multicolumn{3}{l}{mean $F_{1}=57.6$, max $F_{1}=64.0$ reported in~\citet{drozdov2020diora}} \\
\cmidrule(){2-4}
&1 & Our replication using the original codebase\tablefootnote{\url{https://github.com/iesl/s-diora}\label{footnote:sdioracodebase}}($\text{seed}=1943591871$) & $56.3$\\
&2 & Our replication using the original codebase\footnoteagain{footnote:sdioracodebase} ($\text{seed}=0315$)& $60.0$\\
&3 & Our replication using the original codebase\footnoteagain{footnote:sdioracodebase} ($\text{seed}=0075$)& $58.9$\\
&4 & Our replication using the original codebase\footnoteagain{footnote:sdioracodebase} ($\text{seed}=1313$)& $54.7$\\
&5 & Our replication using the original codebase\footnoteagain{footnote:sdioracodebase} ($\text{seed}=442597220$)& $54.9$\\
\hline
\multirow{6}{*}{\rotatebox[origin=c]{90}{ConTest}} &
\multicolumn{3}{l}{mean $F_{1}=62.8$, max $F_{1}=65.9$ reported in~\citet{cao-etal-2020-unsupervised}} \\
\cmidrule(){2-4}
&1 & A checkpoint provided by the authors through personal email & $65.9$\\
&2 & Our replication using the original codebase\tablefootnote{\url{https://github.com/stevenxcao/constituency-test-parser}\label{footnote:ConTestcodebase}} ($\text{id}=0$)& $61.6$\\
&3 & Parsed data provided by the authors through personal email & $62.3$\\
&4 & Our replication using the original codebase\footnoteagain{footnote:ConTestcodebase} ($\text{id}=1$)& $63.0$\\
&5 & Our replication using the original codebase\footnoteagain{footnote:ConTestcodebase} ($\text{id}=2$)& $61.8$\\
\hline
\multirow{7}{*}{\rotatebox[origin=c]{90}{ContexDistort\tablefootnote{Given a pretrained language model, ContexDistort is a deterministic algorithm. Therefore, we used different layers of the language model as runs to obtain different results.}}} &
\multicolumn{3}{l}{$F_{1}=49.0$ reported in~\citet{li-lu-2023-contextual}} \\
\cmidrule(){2-4}
&1 & Our replication using the original codebase\tablefootnote{\url{https://github.com/jxjessieli/contextual-distortion-parser}\label{footnote:ContexDistortcodebase}} on 10th layer of ``bert-base-cased'' & $48.8$\\
&2 & Our replication using the original codebase\footnoteagain{footnote:ContexDistortcodebase} on 12th layer of ``bert-base-cased'' & $46.6$\\
&3 & Our replication using the original codebase\footnoteagain{footnote:ContexDistortcodebase} on 11th layer of ``bert-base-cased'' & $48.7$\\
&4 & Our replication using the original codebase\footnoteagain{footnote:ContexDistortcodebase} on 8th layer of ``bert-base-cased'' & $46.9$\\
&5 & Our replication using the original codebase\footnoteagain{footnote:ContexDistortcodebase} on 9th layer of ``bert-base-cased'' & $48.1$\\
\hline
\end{tabular}
}
\caption{$F_1$ scores are on PTB test for different teachers in different runs. Note that the runs were randomly shuffled for the randomized experiment. }
\label{tab:detailed_variants}
\end{table}

\end{document}